\documentclass{article}
\usepackage{arxiv}
\usepackage[utf8]{inputenc} % allow utf-8 input
\usepackage[T1]{fontenc}    % use 8-bit T1 fonts
\usepackage[colorlinks,
            linkcolor=blue,
            anchorcolor=blue,
            citecolor=green]{hyperref}
\usepackage{url}            % simple URL typesetting
\usepackage{booktabs}       % professional-quality tables
\usepackage{amsfonts}       % blackboard math symbols
\usepackage{nicefrac}       % compact symbols for 1/2, etc.
\usepackage{microtype}      % microtypography
\usepackage{lipsum}		% Can be removed after putting your text content
\usepackage{graphicx, subcaption}
\usepackage{doi}
\usepackage{gensymb}
\usepackage{algorithm}
\usepackage{algpseudocode}
\usepackage{amsmath}
\usepackage{color,xcolor,soul}
\usepackage{siunitx}
\usepackage{enumitem}
\usepackage{natbib}
\usepackage{comment}
\setcitestyle{authoryear,round}
\makeatletter
\newcommand{\customlabel}[2]{%
   \protected@write \@auxout {}{\string \newlabel {#1}{{#2}{\thepage}{#2}{#1}{}} }%
   \hypertarget{#1}{}
}
\makeatother

\title{FengWu: Pushing the Skillful Global Medium-range Weather Forecast beyond 10 Days Lead}

\renewcommand{\shorttitle}{FengWu - Global Medium-range Weather Forecasting}

\author{Kang Chen $^{1,2,*}$ \And
	Tao Han $^{1,*}$ \And
	Junchao Gong $^{1,3,*}$ \And
	Lei Bai $^{1,*,\dagger}$ \And
        Fenghua Ling $^{4}$ \And
        Jing-Jia Luo $^{4}$ \And
        Xi Chen $^{5}$ \And
        Leiming Ma $^{6}$ \And
        Tianning Zhang $^{1}$ \And
        Rui Su $^{1}$ \And
        Yuanzheng Ci $^{1}$ \And
        Bin Li $^{2}$ \And
        Xiaokang Yang $^{3}$ \And
	Wanli Ouyang $^{1}$ \AND
        $^{*}$ \small{Equal Contributions,$^{\dagger}$ Project Lead, bailei@pjlab.org.cn} \AND
        $^{1}$ Shanghai Artificial Intelligence Laboratory \And
        $^{2}$ University of Science and Technology of China \And 
        $^{3}$ Shanghai Jiao Tong University \And
        $^{4}$ Nanjing University of Information Science and Technology \And
        $^{5}$ The Institute of Atmospheric Physics, Chinese Academy of Sciences \And
        $^{6}$ Shanghai Meteorological Bureau
}

\hypersetup{
pdftitle={High-resolution-global-weather-model},
pdfkeywords={Numerical Weather Prediction, Deep Learning},
}

\begin{document}
\maketitle

\begin{abstract}
We present FengWu, an advanced data-driven global medium-range weather forecast system based on Artificial Intelligence (AI). Different from existing data-driven weather forecast methods, FengWu solves the medium-range forecast problem from a multi-modal and multi-task perspective. Specifically, a deep learning architecture equipped with model-specific encoder-decoders and cross-modal fusion Transformer is elaborately designed, which is learned under the supervision of an uncertainty loss to balance the optimization of different predictors in a region-adaptive manner. Besides this, a replay buffer mechanism is introduced to improve medium-range forecast performance. With 39-year data training based on the ERA5 reanalysis, FengWu is able to accurately reproduce the atmospheric dynamics and predict the future land and atmosphere states at 37 vertical levels on a 0.25° latitude-longitude resolution. Hindcasts of 6-hourly weather in 2018 based on ERA5 demonstrate that FengWu performs better than GraphCast in predicting 80\% of the 880 reported predictands, e.g., reducing the root mean square error (RMSE) of 10-day lead global z500 prediction from 733 to 651 $m^{2}/s^2$. In addition, the inference cost of each iteration is merely 600ms on  NVIDIA Tesla A100 hardware. The results suggest that FengWu can significantly improve the forecast skill and extend the skillful global medium-range weather forecast out to 10.75 days lead (with ACC of z500 > 0.6) for the first time.
\end{abstract}


\section{Introduction}\label{sec:introduction}
%\bai{Significance and value of NWP} 
Understanding and predicting the Earth's environment we live on, especially the atmosphere system, is a long-standing pursuit of human beings. Meteorological phenomena were recognized 3,000 years ago in the inscriptions on borns or tortoise shells of the Shang Dynasty in China (ca. sixteenth to eleventh century BCE)~\citep{Di2008}. The atmosphere system attracts more attention under global warming and the upsurge of extreme weather events ~\citep{lehmann2015increased,rahmstorf2011increase}.
Weather forecast, which involves analysis of past and present weather observations to predict future atmospheric conditions ranging from hours to days and even weeks, plays a critical role in routine decision-making for agriculture management, transportation, natural disasters (e.g., floods and storms) prevention, and green energy production, etc. 
For millennia, people have recognized the importance of weather forecast and have sought diverse ways to predict the weather. For example, the ancient Babylonians attempted to predict the weather based on sky observations~\citep{taub2004ancient} and the ancient Chinese developed instruments made of copper or wood to measure the wind speed.

Among various weather prediction tasks, global medium-range weather forecast, which targets predicting future global atmospheric conditions up to fourteen days ahead~\citep{bengtsson1985medium}, is arguably one of the most highly demanded tasks. It not only serves as the foundation for the deployed global ensemble forecast system~\citep{gneiting2005weather} to directly provide weather forecast services, but also provides background information and boundary conditions for regional numerical weather forecast systems~\citep{mass1998regional}. 
%\bai{one sentence about early global MWF before 1980s.}
Electronic computers were utilized in the 1950s for medium-range weather forecast by solving the partial differential equations ~\citep{bolin1955numerical}. Still, they were not widely available due to limitations in computing resources and data availability ~\citep{lynch2008origins}. However, with the rapid development of Earth observation techniques (e.g., satellites) and High-Performance Computing (HPC) facilities, an increasing number of physical processes (e.g., radiation, thermodynamics, and fluid dynamics) could be simulated in a higher resolution with more accurate observations, leading to apparent forecast skills improvements from the 1980s.

Despite the significant breakthroughs achieved in the past decades, the performance of global medium-range weather forecast systems is still limited in forecast accuracy and extendibility due to large uncertainties in initial and boundary conditions, complicated non-linear physical processes, and heavy computation costs ~\citep{bauer2015quiet}. With the accumulation of massive weather observations and the maturity of deep learning techniques (e.g., large-scale training frameworks), researchers have commenced exploration on the possibility of AI-driven Numerical Weather Prediction (NWP) models. 
Specifically, Rasp et al. ~\citep{rasp2020weatherbench} first introduced ResNet ~\citep{he2016deep} to generate 5.625$^{\circ}$ latitude-longitude resolution of global weathe prediction. Hu ~\citep{hu2023swinvrnn} utilized Recurrent Neural Network (RNN) architecture with variational loss to improve long-lead forecasts. While these attempts reveal the potential of data-driven methods in numerical weather prediction, they are limited in low-resolution data, leading to limited forecast applications. 
Recently, FourCastNet ~\citep{pathak2022fourcastnet}, the first model producing 0.25$^{\circ}$ resolution forecasts, applies Vision Transformer(ViT) ~\citep{dosovitskiy2020image} and Adaptive Fourier Neural Operators (AFNO)~\citep{guibas2021adaptive} for efficient computation. 
Then, PanGu ~\citep{bi2022pangu} acquires promising medium-range performance that surpasses ECMWF Integrated Forecasting System (IFS) in 0.25$^{\circ}$ resolution with a multi-timescale model combination strategy based on four 3D Earth-Specific Transformers. 
GraphCast~\citep{lam2022graphcast} further boosts the AI methods' upper bound in NWP. It is more accurate in predicting 90\% of the atmospheric variables compared with the ECMWF's deterministic operational forecasting system(IFS-HRES). In GraphCast, Graph neural network (GNN) is employed for medium-range global weather forecast, a 12-step autoregressive (AR) finetuning is adapted as the strategy for increasing the long-lead prediction accuracy, and mean squared error (MSE) loss used in GraphCast is elaborately weighted based on the pressure levels and weather variables.

\begin{figure}[htbp]
\centering

\begin{minipage}[b]{0.49\linewidth}
\centering
\includegraphics[width=\textwidth]{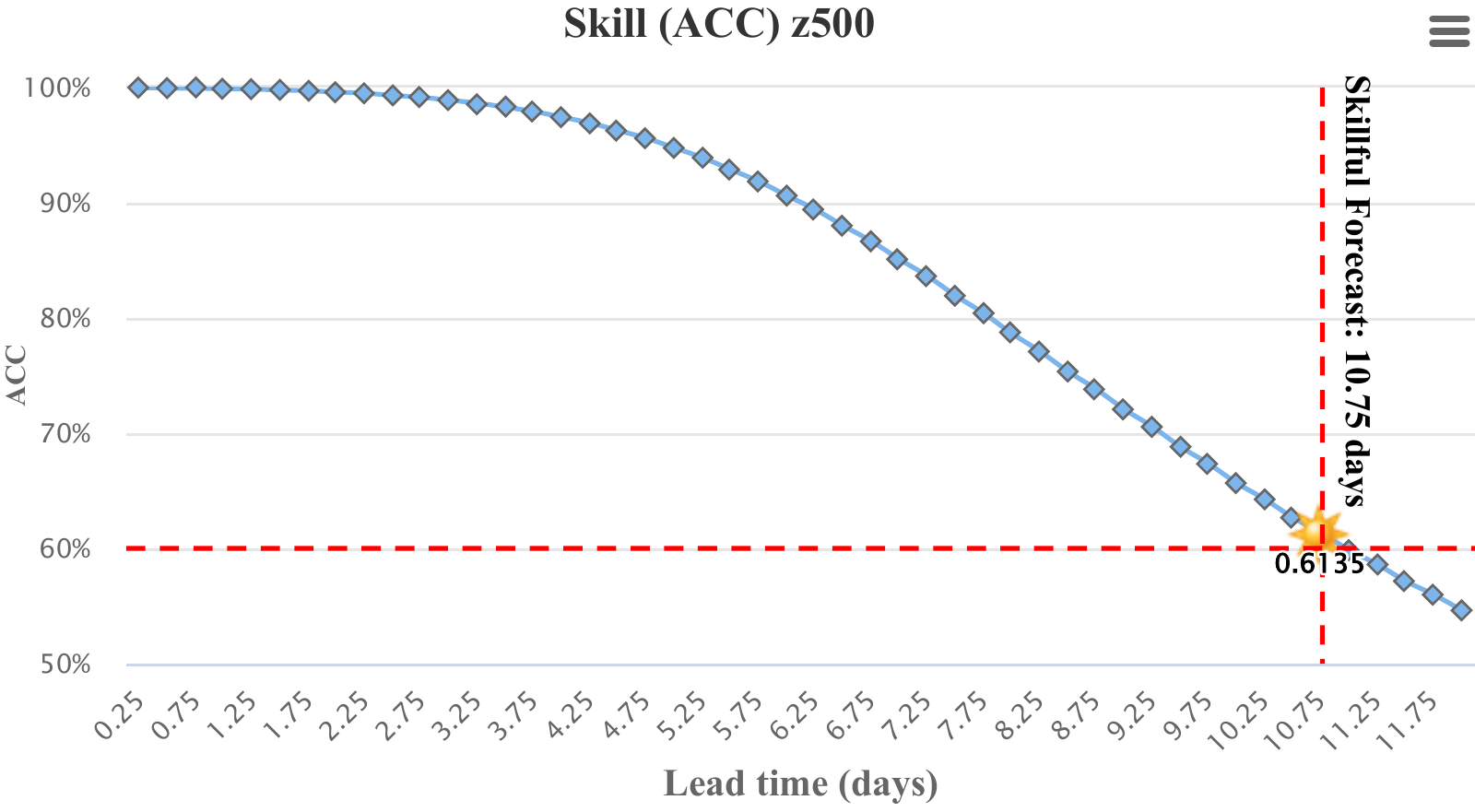}
\end{minipage}
\hfill
\begin{minipage}[b]{0.49\linewidth}
\centering
\includegraphics[width=\textwidth]{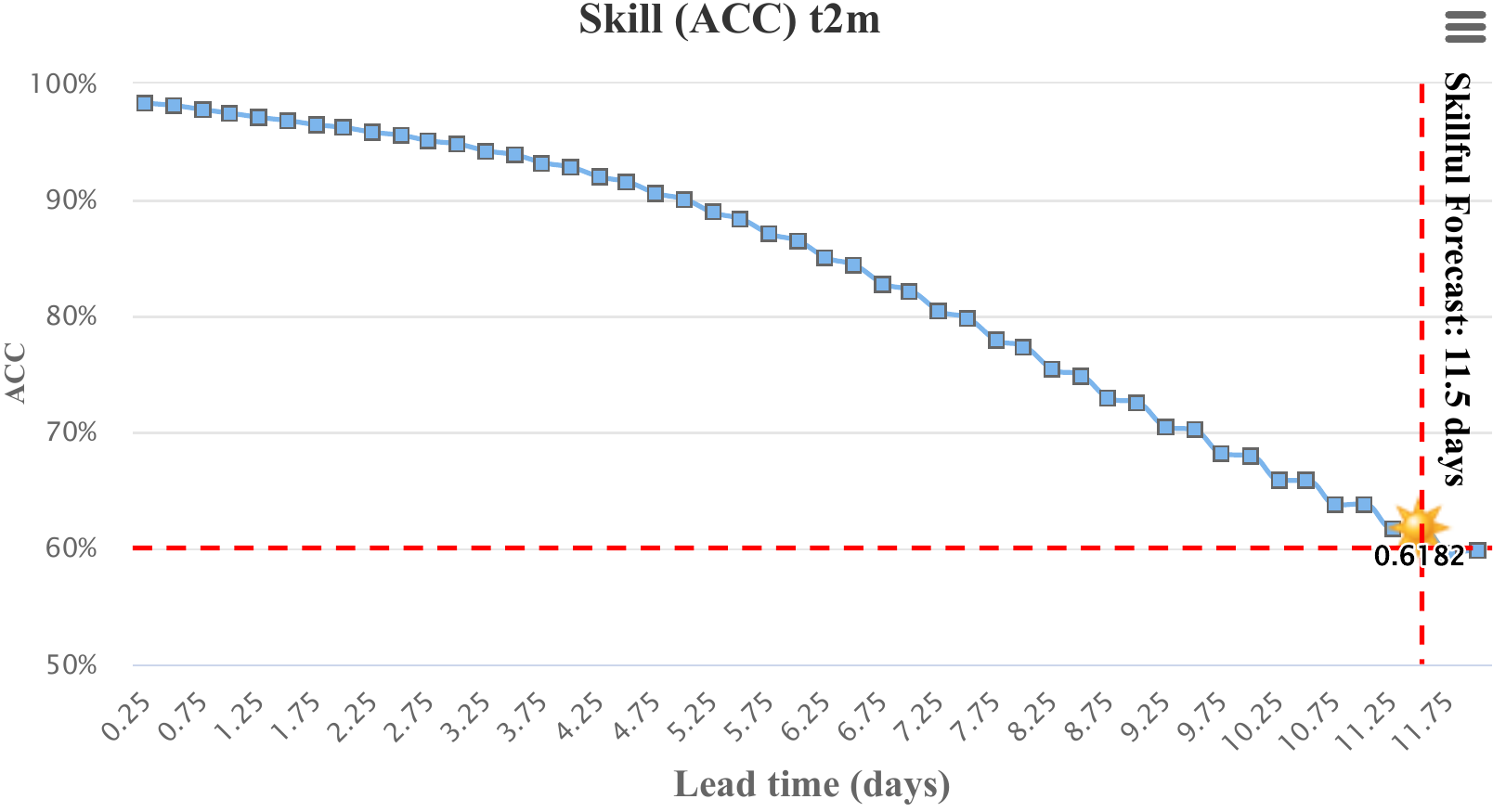}
\end{minipage}

\caption{\textbf{The skillful forecast lead time of FengWu on z500 and t2m.} With the ACC > 0.6 as the criterion of a skillful weather prediciton system, our findings are that FengWu can push the skillful forecast lead times to 10.75 days and 11.5 days for z500 and t2m, respectively.}
\end{figure}

In this paper, we propose an AI model to improve the medium-range weather forecast from a multi-modal and multi-task perspective. 
Specifically, our first proposition is to recognize the high-dimensional weather data, consisting of different atmosphere variables(e.g., temperature, winds, geopotential height, etc.) in differentpressure levels, as distinct modalities, with each variable being treated as a modality; this is different from existing AI-based solutions that stack all variables as single-modal input. 
This approach enables us to leverage existing multi-modal designs, e.g. processing the data of different atmosphere factors with modal-specific encoders, followed by a cross-modal Transformer to model the interactions among all atmosphere variables. The future states of each variable is derived separately after the cross-modal Transformer with modal-specific decoders. 
Our second proposition is to highlight that NWP is a multi-task regression problem based on the view that the prediction of each variable can be treated as a unique task. %Existing weather forecast methods use a na\"ive weighted sum of losses, where the loss weights are uniform or manually tuned~\cite{}. However, 
The prediction of some atmosphere variable will be more difficult than others, making the methods that treat different tasks equally unable to reach the global minimum point. Manually tuning the optimal weight for each task (variable) is prohibitively expensive and difficult. Treating NWP as a multi-task problem, we can leverage the uncertainty loss developed in the multi-task learning paradigm to the weather forecast domain, which considers the homoscedastic uncertainty of each variable. By learning to automatically scale weights of variable regression, introducing multi-task learning can improve learning effectiveness, leading to better weather prediction accuracy.

Another issue in AI-driven global medium-range weather forecast is generating long-lead predictions (e.g., 10 days' predictions), which is hard to be directly optimized due to the extremely huge volume of global weather data (e.g., 2.3 GB for each time step in the ERA5 pressure level dataset with float32 format) even with the most advanced GPU devices. Previous works tackled this problem either by autoregressively fine-tuning with two~\citep{pathak2022fourcastnet} or multiple steps~\citep{lam2022graphcast} powered by elaborate engineering techniques, or training separate models for different steps~\citep{bi2022pangu}. Despite they are demonstrated to be effective, these solutions can be computationally expensive and memory-intensive, making them challenging to implement when computation resources are limited. 
To solve the long-lead prediction issue, we propose the Replay Buffer mechanism, which is inspired by the reinforcement learning study ~\citep{schaul2015prioritized}. The replay buffer stores the predicted results from previous optimization iterations and uses them as the current model's input, which mimics the intermediate input error during the auto-regressive inference stage. This design boosts the long-lead forecast quality with efficient computation and memory.

Based on the above perspectives and designs, we develop FengWu \footnote{The name of ''FengWu'' comes from the ancient Chinese anemometer used from the Han dynasty, considered as the earliest prototype for measuring wind speed and orientation.}, an advanced weather forecast system for global medium-range weather predictions.
By training with the high-resolution (i.e., 0.25° latitude-longitude resolution) ERA5 dataset over the past 39 years, FengWu is able to accurately emulate the atmospheric dynamics and predict the future land and atmosphere status of 37 levels.
Hindcasts of 6-hourly weather in the year 2018 indicate that FengWu achieves the best forecast skills among all the released data-driven forecast systems. 
Specifically, FengWu has higher accuracy than GraphCast on 80\% of the 880 reported predictands.
%and IFS-HERS (i.e., the most advanced dynamical model forecast system developed by ECWMF), separately. 
The forecast skill improvements make the skillful forecast lead time of FengWu reach 10.75 days (ACC of z500 $>$ 0.6 \cite{bauer2015quiet}) for the first time by an AI-based approach.

\section{Preliminary} \label{sec:preliminary}
\subsection{Dataset} 
ERA5~\citep{hersbach2020era5} is a global atmospheric reanalysis dataset produced by the European Centre for Medium-Range Weather Forecasts (ECMWF). It provides comprehensive information about the Earth's climate and weather conditions, covering the period from 1940 to the present. ERA5 provides a wide range of variables such as temperature, humidity, precipitation, wind speed and direction, mean sea level pressure, and many others. The data is available at a high spatial resolution of 0.25° latitude-longitude resolution and 37 vertical pressure levels, ranging from 1000 hPa to 1 hPa, which makes it suitable for a wide range of applications, including climate research, weather forecast, and environmental monitoring. 

In this study, FengWu simulates 5 atmospheric variables (each with 37 pressure levels) and four surface variables, a total of 189 predictands. Specifically, the atmospheric variables are geopotential ($z$), relative humidity ($r$),  zonal component of wind ($u$), meridional component of wind ($v$), and air temperature ($t$), whose 37 sub-variables at different vertical level are presented by abbreviating their short name and pressure levels (e.g., z500 denotes the geopotential height at a pressure level of 500 hPa). And the four surface variables are 2-meter temperature (t2m), 10-meter u wind component (u10), 10-meter v wind component (v10), and mean sea level pressure (msl). 

For consistency, we follow the validation strategies demonstrated by GraphCast, i.e., the data from 1979-2015 is used for training, 2016-2017 for validation, and 2018 for testing. In addition, we also leverage the 6-hourly sampled data (T00, T06, T12, T18) instead of the hourly ERA5 dataset for training. 

\subsection{Problem Formulation}
 
The objective of FengWu is to utilize AI techniques to build the most skillful high-resolution medium-range global weather forecast system, which predicts the future 14-day global atmosphere states based on the current atmosphere conditions. 
%To resolve it, we use a two-step approach following GraphCast, as illustrated in Figure xx. In detail, 
Formally, we denote the weather state at time slot $i$ as a high dimension tensor $X^{i}\in \mathbb{R}^{C \times W \times H}$, where $C$ denotes the number of atmosphere variables considered in this work, $W$ and $H$ are the width and height. When mapping the continuous atmosphere fluid to a 2D spatial plane with 0.25° latitude-longitude resolution, we have $C=189, W=721,$ and $H=1440$. FengWu aims at generating 14-day lead time forecasts $\{\hat{X}^{i+1},\hat{X}^{i+2},...,\hat{X}^{i+56} \}$ with a time-interval of six hours,
\begin{equation} \label{eq:formualtion}
\{\hat{X}^{i+1},\hat{X}^{i+2},...,\hat{X}^{i+56} \} = \operatorname{FengWu}\left(X^{i}\right),
\end{equation}
where $\hat{X}^{i+\tau}$ is the prediction of the weather state at time slot $i+\tau$. However, it is hard to directly learn a function for Eq.~\ref{eq:formualtion} due to the extremely large size of the global atmosphere data. Following the practice of atmosphere simulation, FengWu targets on learning a function to predict the data of the next step, which could then generate multi-step predictions in an auto-regressive manner, i.e., 

\begin{equation} \label{eq:reformualtion}
\hat{X}^{i+1}=\operatorname{FengWu}\left(X^{i}\right), \ \hat{X}^{i+2}=\operatorname{FengWu}\left(\hat{X}^{i+1}\right), \
..., \
\hat{X}^{i+56}=\operatorname{FengWu}\left(\hat{X}^{i+55}\right)
\end{equation}

\section{Method} \label{sec:method}
In this section, we introduce the details of FengWu, which includes three main components: 1) the network Architecture including Transformer-based modal-customized encoder-decoders and cross-modal fuser, 2) the uncertainty loss for multi-task optimization, and 3) the replay buffer for long-lead predictions.

\subsection{Network Architecture}
%\bai{justification and design of multi-modal network structure}

\begin{figure}
  \centering      \includegraphics[width=0.95\textwidth]{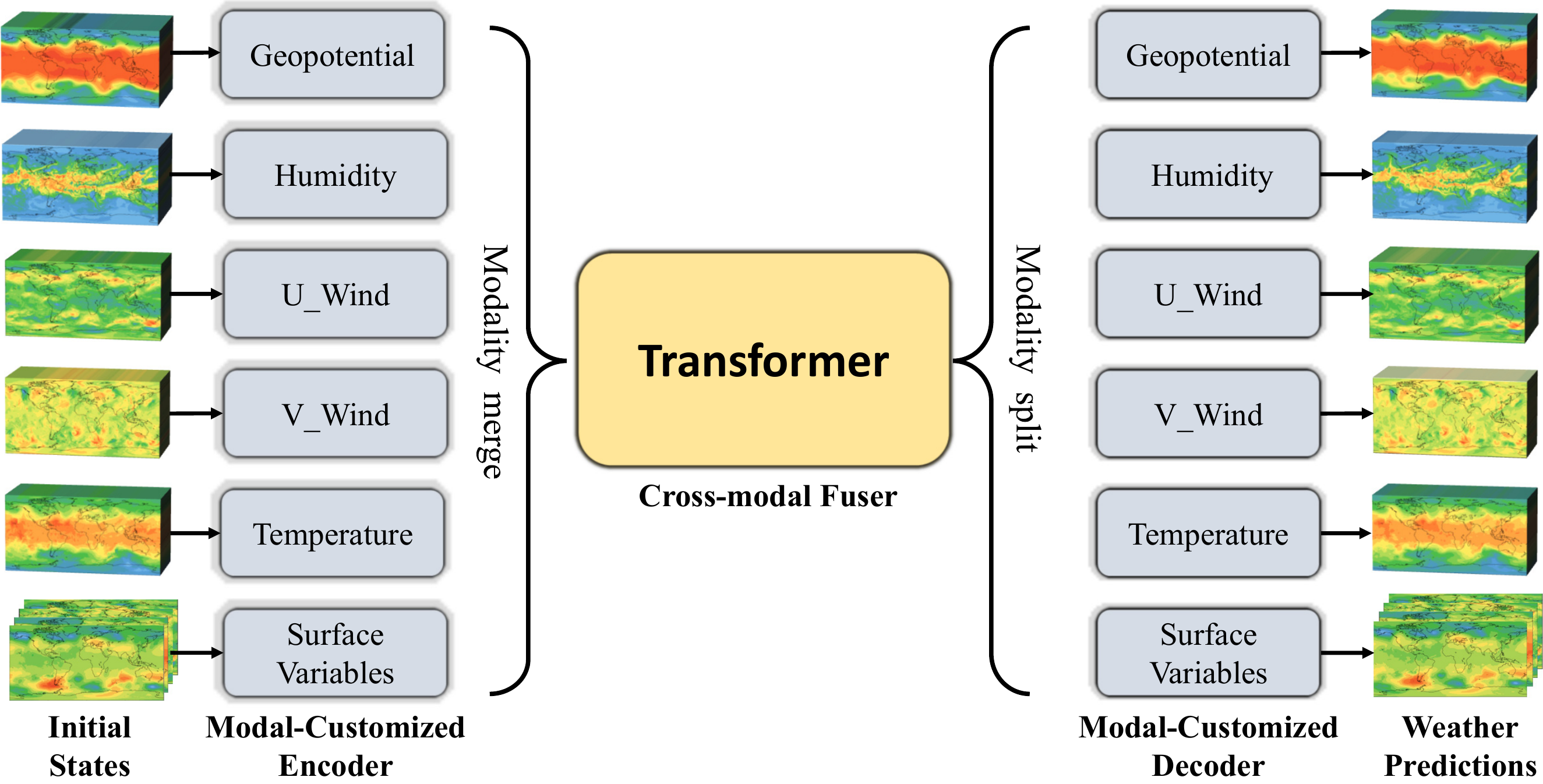}
    \caption{Overview of FengWu's architecture. FengWu first treats the multiple weather factors as different modalities and extracts their feature embeddings independently. And then a transformer-based network is utilized to fuse and pass messages among different modalities. Finally, the high-level feature representation is used to get the predictors via the modal-customized decoder.}
    \label{fig:net}
\end{figure}

FengWu considers weather variables as different modalities of the atmosphere state and employs an "encode-fuse-decode" structure as illustrated in Figure~\ref{fig:net}. The \emph{Modal-Customized Encoder} encodes multi-modal features, which are fused by a Transformer-based \emph{Cross-modal Fuser} to get the joint representations. The \emph{Modal-Customized Decoder} then separately predicts the weather variables from the joint representations.

\paragraph{Modal-Customized Encoder.} We design Modal-Customized Encoders to extract features independently. 
Each weather state $X^{i}$ with shape $(C, W, H)$ is sliced into earth surface state $X^{i}_{s}$, 
geopotential state $X^{i}_{z}$,
humidity state $X^{i}_{q}$, the eastward component of the wind state $X^{i}_{u}$,  the northward
component of wind state $X^{i}_v$, and temperature state $X^{i}_t$ whose shapes are, respectively, $(C_{s}, W, H)$, $(C_{z}, W, H)$, $(C_{q}, W, H)$, $(C_{u}, W, H)$, $(C_{v}, W, H)$, and $(C_{t}, W, H)$. 
To obtain features $\Tilde{X}_{m}$ for $m \in \{s, z, q, u, v, t\}$ separately, a transformer-based encoder $f_{en, m}({X}_{m}|\theta_{en, m})$ with encoder parameters $\theta_{en, m}$ is used for its corresponding state ${X}_{m}$. The output of the encoder is denoted as $Z_{m}$ for the modality $m$, where 
\begin{equation}
Z_{m} = f_{en, m}({X}_{m}|\theta_{en, m}).
\end{equation}

\paragraph{Cross-modal Fuser.} 
The output of the encoder $Z_{m}$ for $m \in \{s, z, q, u, v, t\}$ are concatenated to obtain the fused features as follows:
\begin{equation}
Z = concat(Z_s, Z_z, Z_q, Z_u, Z_v, Z_t),
\end{equation}
where $concat$ denotes feature concatenation along the feature channel dimension. The fused features are then fed into a transformer for fusing their information and extracting the fused features $\tilde{Z}$.

\paragraph{Modal-Customized Decoder.} In Multi-task Decoder, tokens generated by  Multi-modal Feature Fuser are used for predicting the mean and variance of atmosphere variables. 
Separate modality decoders $f_{de, m}(\tilde{Z} | \theta_{de, m})$ for $m \in \{s, z, q, u, v, t\}$ are designed to predict the future state of corresponding modalities, where $\theta_{de, m}$ denotes the parameters of the decoder $f_{de, m}$ for modality $m$. The decoder network is similar to the encoder.

\subsection{Uncertainty Loss for Multi-task Optimization}
In this paper, weather forecast learning is regarded as multi-task learning.  
Previous studies in the multi-task learning paradigm show that assigning different weights between different tasks is helpful for learning representations.
This observation is consistent with the practice in GraphCast ~\citep{lam2022graphcast}, which assigns manually designed weights for different variables and pressure levels as hyperparameters. While it is demonstrated to be effective, manually determining approximate weights for variables would be arduous and sub-optimal. 
 
To solve the multi-task optimization issue more accurately and elegantly, we introduce the uncertainty loss to automatically learn weights for weather forecasting. Specifically, FengWu is defined as a probabilistic model that predicts the parameters $\hat{\mu}^{i+1}, \hat{\sigma}^{i+1}$ of a Gaussian distribution:
\begin{align}
\hat{\mu}^{i+1}, \hat{\sigma}^{i+1} = \operatorname{FengWu}\left(X^{i}\right)
\label{eq:probability_model}
\end{align}
where $\hat{\mu}^{i+1}$ and $\hat{\sigma}^{t+1}$ are predicted mean and variance of predictands $X _{i+1}$. And the probability of atmosphere variables can be calculated by the mean and variance:
\begin{align}
p\left(x^{i+1}_{c, w, h} \mid \hat{\mu}^{i+1}, 
\hat{\sigma}^{i+1}
\right)=\mathcal{N}
\left(
\hat{\mu}^{i+1}_{c, w, h}, \hat{\sigma}^{i+1}_{c, w, h}
\right)
\label{eq:likelihood}
\end{align}
In Eq.~\ref{eq:likelihood}, each element $x^{i+1}_{c, w, h}$ in $X^{i+1}$  with  subscript $(c, w, h)$ follows an independent univariate Gaussian distribution $\mathcal{N}
\left(
\hat{\mu}^{i+1}_{c, w, h}, \hat{\sigma}^{i+1}_{c, w, h}
\right)$, where $c= 1, \ldots, 189$ denotes the index for the channel, i.e. different pressure levels and weather variables, e.g. temperature. $w$ and $h$ respectively denote the latitude grid and longitude grid. 
We adopt maximum likelihood estimation to allocate weights for different tasks(variables). 
Because we employ the likelihood as the minimization objective, the loss of each variable $c$ at location $(w, h)$ is automatically weighted by homoscedastic uncertainty. The uncertainty loss provides an approach to tradeoff the weights between variables, pressure levels, and locations without an expensive manual search, which is efficient, particularly in large weather forecast models.

\subsection{Replay Buffer for Long-lead Predictions}
\label{Sec:relpay_buffer}

Trained as a single-step predictor, directly using FengWu for medium-range prediction will lead to inferior long-lead forecast performance due to the accumulation of errors in the AutoRegressive (AR) inference process. GraphCast effectively eases this problem by adding an autoregressive training stage, which gradually increases the number of autoregressive steps in the training from 2 to 12.
However, this approach encounters two challenges. Firstly, the intermediate predictions generated during multiple forwards can be wasteful, as they are discarded after being optimized and are not reused for other autoregressive steps, which reduces their efficiency and slows down the training process. Secondly, the memory required to store and process the gradient in the constantly increasing autoregressive steps can become excessive, limiting the maximum AR steps that can be processed.  

To tackle the problems mentioned above, this work proposes a replay buffer mechanism.  
Specifically, a set $\mathcal{B}=\{\hat{X}^{i+\tau}_{j}\}_{j=0}^{N}$ containing $N$ predictions are denoted to represent the data in the buffer.
%where $\tau\in[1,12]$ is the autoregressive step. 
Initially, the replay buffer first pushes a certain number of first-step predictions in the initial stage. 
In the next stage, FengWu learns from both the original dataset and the replay buffer, e.g., sampling data from both the original dataset and the replay buffer. Accordingly, the predicted results, either taking input from the original data or the replay buffer, are treated as the intermediate predictions and saved to the data buffer, resulting in diverse finetuning frequencies between different autoregressive steps. 
The last element in the replay buffer will be popped out if the queue is full for each data collection. 
The replay buffer plays a critical role in enabling our system to perform long-lead autoregressive forecasts by collecting and reusing intermediate predictions, thus enforcing the FengWu to take the accumulated autoregressive estimation errors into consideration during training. 
This online learning strategy is particularly valuable in situations where the devices or framework does not support long-lead AR training. 
Specifically, a training sample at time step $i+1$ stored in the replay buffer is used as the input of FengWu for prediction at time step $i+2$ and its predicted results at time step $i+2$ are stored in the replay buffer, and the predicted results at time step $i+2$ in the replay buffer will be used as the input in the latter training iteration.
Therefore, the replay buffer helps the training stage to simulate the long-lead autoregressive forecast during the inference in Eq.~\ref{eq:reformualtion}. 
In addition to its role in enabling long-lead AR forecast, the replay buffer also offers the benefit of reducing GPU memory usage by storing data on the CPU. Overall, the replay buffer enhances the efficiency and effectiveness of the long-lead AR learning process.

\begin{figure}[ht!]
  \centering      \includegraphics[width=0.999\textwidth]{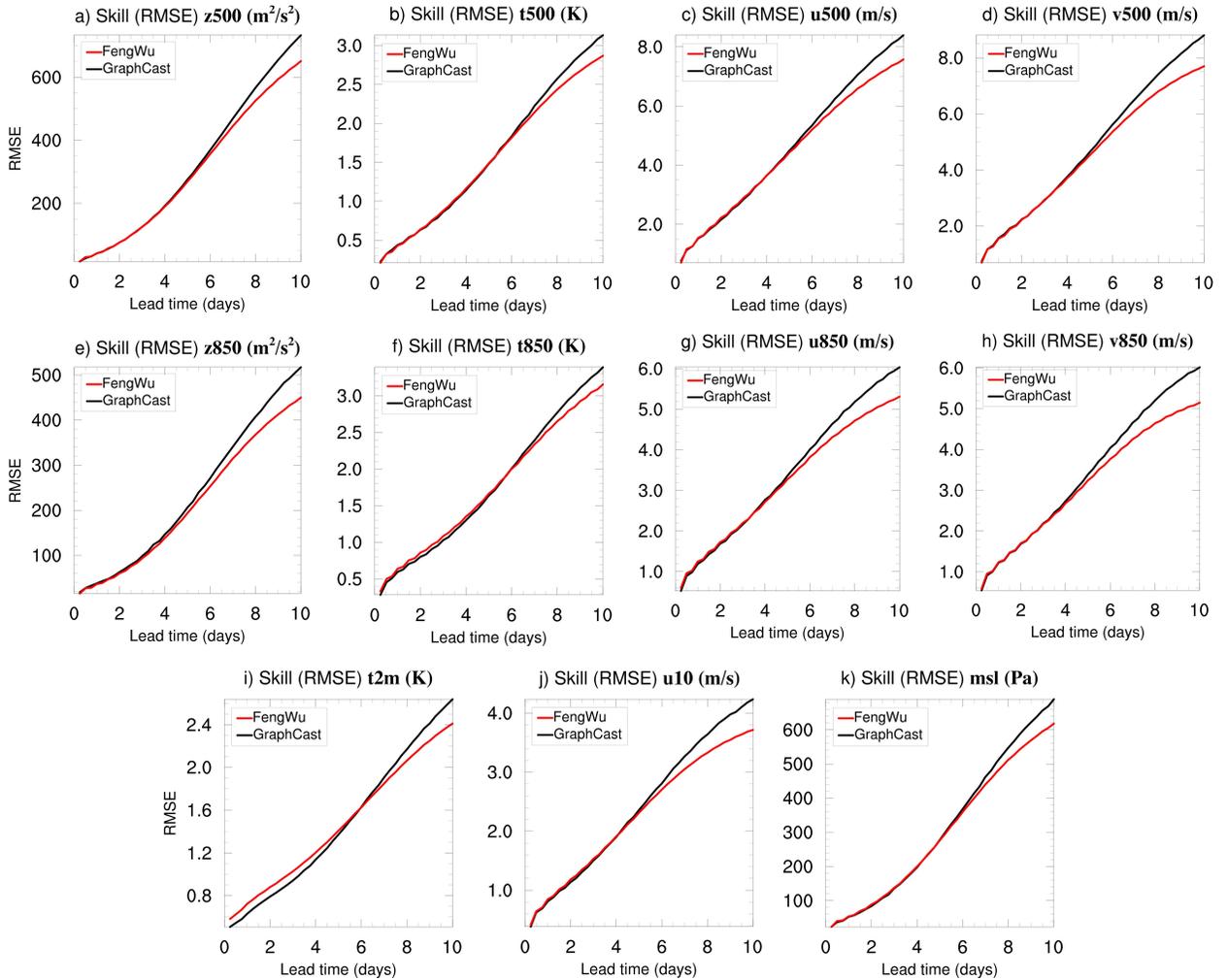}
    \caption{\textbf{
     Latitude-weighted RMSE skill of FengWu (\textcolor{red}{red} lines) and GraphCast (black lines) predicting the weather in 2018 (Lower RMSE is better).} The x-axis in each sub-figure represents lead time, at a 6-hour interval over a 10-day lead time. The y-axis represents the latitude-weighted RMSE defined in Eq.~\ref{eq:MSE}.}
    \label{fig:result_RMSE}
\end{figure}

\section{Results} \label{sec:results}

\subsection{Evaluation Strategies}
For consistency, we follow the evaluation protocols implemented in the work by GraphCast, which includes the same evaluation metrics, dataset splitting, and lead time of forecast.  
With $00z$ and $12z$ as the initial weather states for each day, we compare the performance of FengWu and GraphCast for a 10-day forecast using the commonly used RMSE and ACC metrics based on the test set.

\begin{figure}
  \centering      \includegraphics[width=0.999\textwidth]{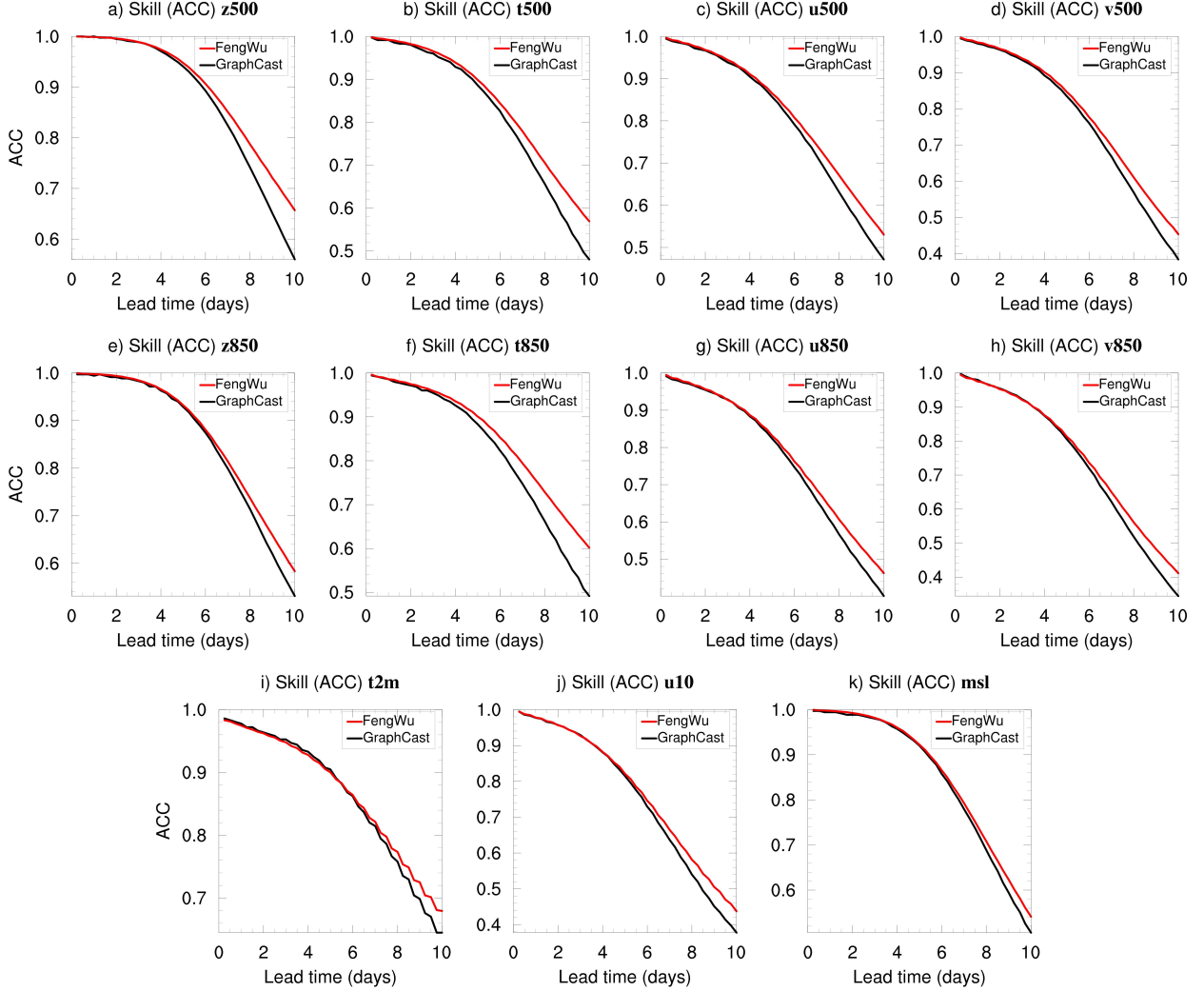}
    \caption{\textbf{ACC skill of FengWu and GraphCast predicting the weather in 2018 (Higher ACC is better).} The x-axis in each sub-figure represents lead time, at a 6-hour interval over a 10-day lead time. The y-axis represents the ACC defined in Eq.~\ref{eq:ACC}.}
    \label{fig:result_ACC}
\end{figure}

\textbf{RMSE} represents the latitude-Weighted Root Mean Square Error, which is a statistical metric widely used in geospatial analysis and climate science to assess the accuracy of a model's predictions or estimates of temperature, precipitation, or other meteorological variables across different latitudes. Given the prediction result $\hat{x}_{c, w, h}^{i+\tau}$ and its target (ground truth) $x_{c, w, h}^{i+\tau}$, the RMSE is defined as follows:

\begin{equation}
\label{eq:MSE}
\operatorname{RMSE}(c,\tau ) =\frac{1}{T} \sum_{i=1}^{T} \sqrt{\frac{1}{W\cdot H}\sum_{w=1}^W\sum_{h=1}^H W \cdot \frac{\operatorname{cos}(\alpha_{w,h})}{\sum_{w'=1}^W \operatorname{cos}(\alpha_{w',h})}(x_{c,w,h}^{i+\tau} - \hat{x}_{c,w,h}^{i+\tau})^{2}},
\end{equation}
where $c$ denotes the index for channels that are either the surface variable or the atmosphere variable at a certain pressure level.  %$C=189$ is the number of prediction targets. 
$w$ and $h$ respectively denote the indices for each grid along the latitude and longitude indices. $\alpha_{w,h}$ is the latitude of point $(w,h)$. $T$ is the total number of valid test time slots.

\textbf{ACC} is the Latitude-weighted Anomaly Correlation Coefficient that evaluates the performance of dynamical models by comparing their predictions of anomalies (departures from the long-term averaged climatology) to observed anomalies. ACC is similar to the standard Anomaly Correlation Coefficient, but also with a latitude weighting factor applied to account for the varying area represented by different latitudes on a spherical Earth,

\begin{equation}
\label{eq:ACC}
\operatorname{ACC}(c,\tau ) =\frac{1}{T} \sum_{i=1}^{T} \frac{\sum_{w,h} W \cdot \frac{\operatorname{cos}(\alpha_{w,h})}{\sum_{w'=1}^W \operatorname{cos}(\alpha_{w',h})} (x_{c,w,h}^{i+\tau} - C_{c,w,h}^{i+\tau}) (\hat{x}_{c,w,h}^{i+\tau} - C_{c,w,h}^{i+\tau})}{
\sqrt{\sum_{w,h}W \cdot \frac{\operatorname{cos}(\alpha_{w, h})}{\sum_{w'=1}^W \operatorname{cos}(\alpha_{w',h})} (x_{c,w,h}^{i+\tau} - C_{c,w,h}^{i+\tau})^{2} \sum_{w,h} W \cdot \frac{\operatorname{cos}(\alpha_{w,h})}{\sum_{w'=1}^W \operatorname{cos}(\alpha_{w',h})} (\hat{x}_{c,w,h}^{i+\tau} - C_{c,w,h}^{i+\tau})^{2}}
},
\end{equation}
where $C_{c,w,h}^{i+\tau}$ is the climatological mean over  the day-of-year containing the validity time $i+\tau$ for a given weather variable $c$ at longitude $w$ and latitude  $h$. It is averaged from the years 1993 to 2016 with the ERA5 data on a daily basis, which is consistent with the approach taken by GraphCast. Note that if $C_{c,w,h}^{i+\tau}$ is calculated hourly, the resulting ACC metric may be significantly higher than the values obtained using the daily climate mean. This is because the ACC metric is sensitive to outliers and can be influenced by various factors, such as the choice of the time period and spatial resolution of the data.

\subsection{Quantitative Skill Evaluation}

Figures \ref{fig:result_RMSE} and \ref{fig:result_ACC} illustrate the predictive performance comparison of FengWu (red lines) and GraphCast (black lines) in terms of RMSE and ACC, respectively. The analysis is conducted over 880 targets at 6-hour intervals, with a lead time of 10 days. The results show that FengWu has both lower RMSE and higher ACC than GraphCast on 80$\%$ of the targets analyzed. 
FengWu demonstrates a comparable level of forecasting accuracy as GraphCast for predicted variables within a lead time range of 1$\sim$5 days, except for t2m. 
In particular, as the lead time increases,  significant improvement with FengWu is observed for all predicted variables, demonstrating FengWu's remarkable ability for long-lead  weather forecasting.

\subsection{Qualitative Prediction Evaluation}

\begin{figure}[ht!]
  \centering      \includegraphics[width=0.99\textwidth]{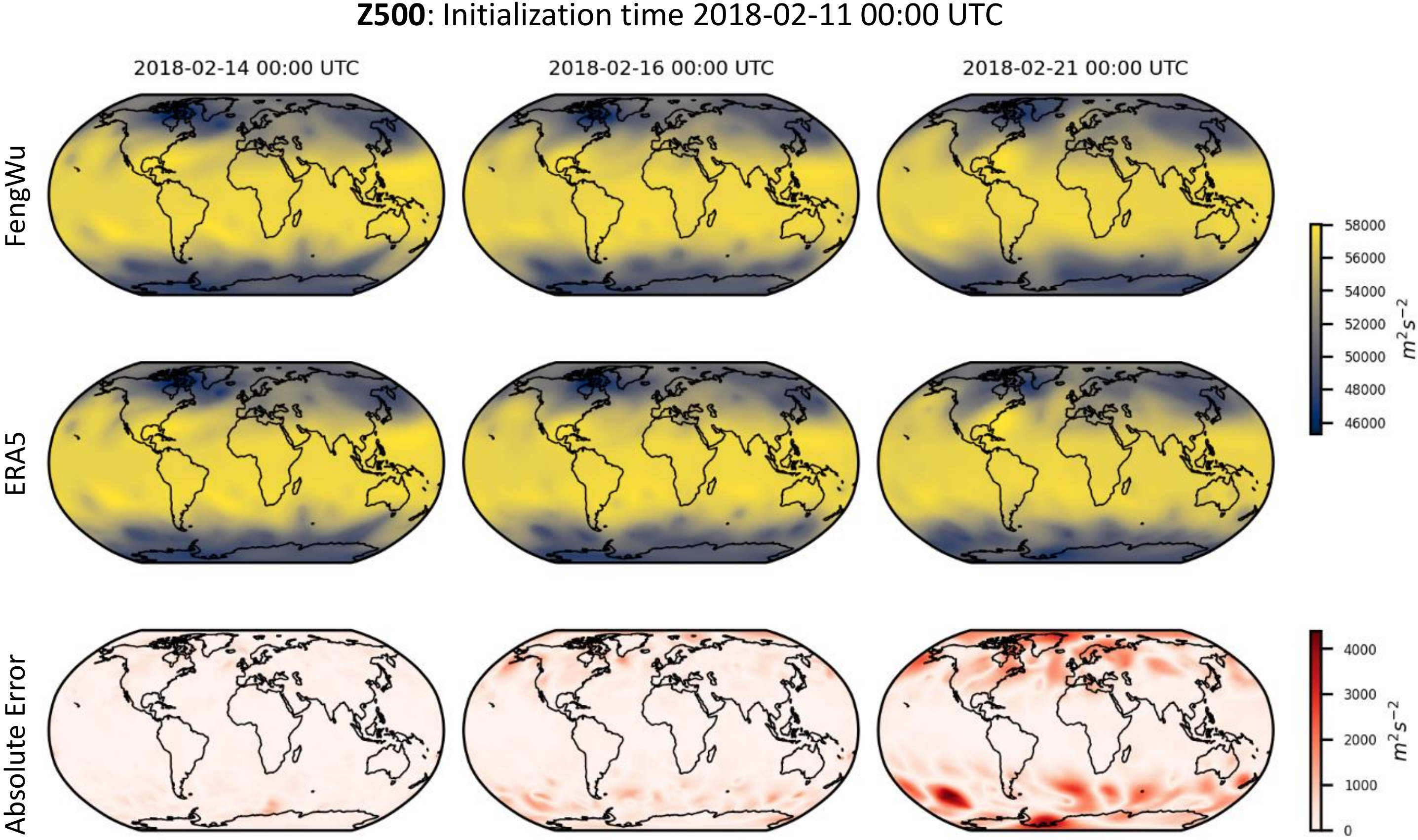}
    \caption{\textbf{Forecast images and absolute error for z500}. Figures of z500 on days 3, 5, and 10 are presented with initialization time at 2018-02-11 00:00 UTC. The subtitles at the top of the columns indicate the dates of prediction. The first row and second row show FengWu and ERA5 ground truth, respectively. Row 3 shows the absolute error between FengWu and ERA5. }
    \label{fig:z500}
\end{figure}

\begin{figure}[ht!]
  \centering      \includegraphics[width=0.99\textwidth]{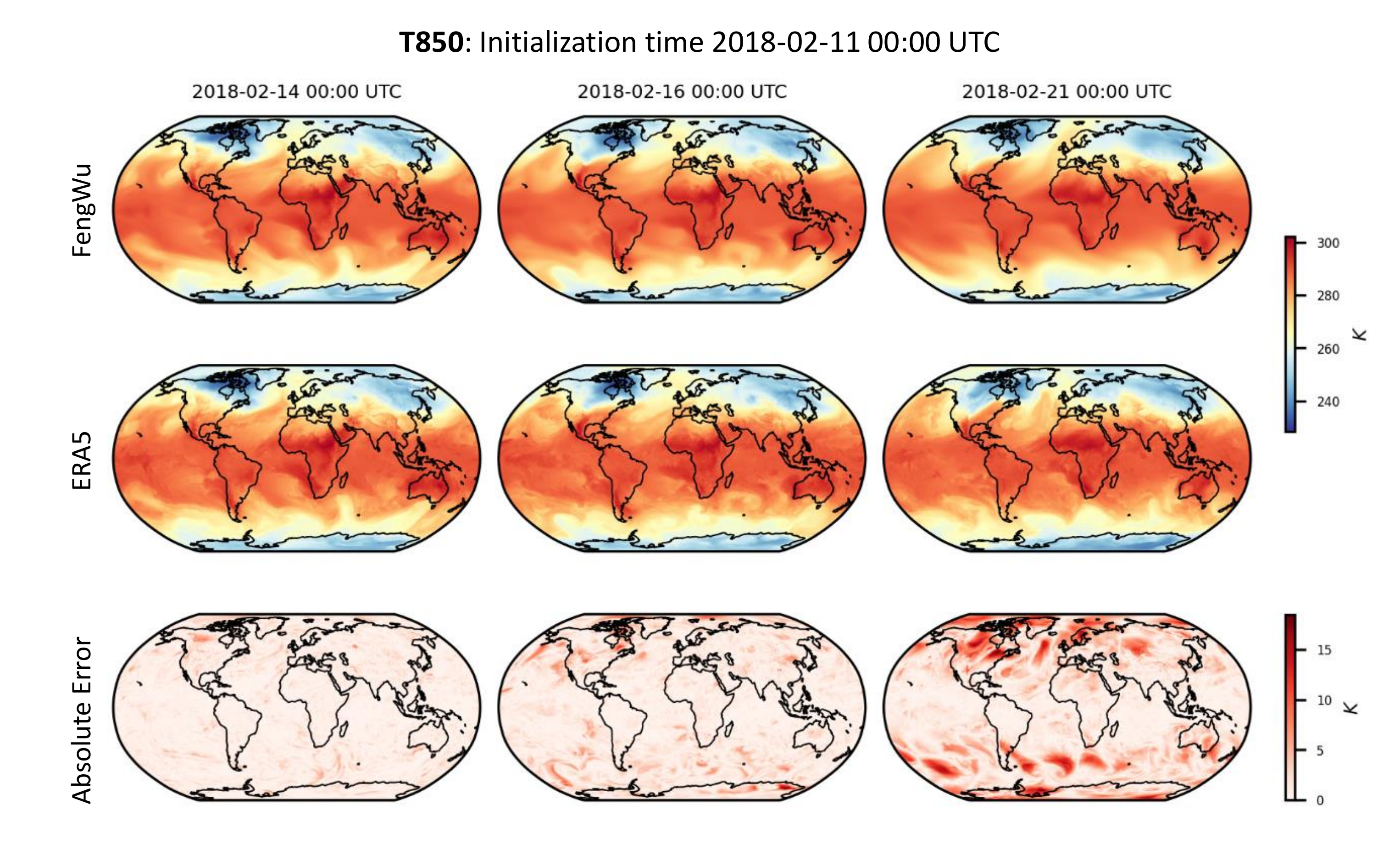}
    \caption{\textbf{Forecast images and absolute error for T850}. The plot demonstrates ERA5 ground truth and FengWu's prediction for T850, with forecast initialization 2018-02-11 00:00. Other settings are similar to Figure~\ref{fig:z500}. }
    \label{fig:t2m}
\end{figure}

We visualize the predicted results of FengWu at lead days 3, 5, and 10 for two variables, i.e., z500 (geopotential at the pressure level of 500 hPa) and t850 (the temperature at the pressure level of 850 hPa), and compare the predictions with the ERA5. In Figure~\ref{fig:z500} and Figure~\ref{fig:t2m}, the top two rows show the sequences of states from FengWu and ERA5, and the third row shows the absolute value of the error from FengWu to ERA5. In both visualizations, FengWu has outcomes close to ERA5 on the third day. As the forecast step increases, the absolute error increases and diffuses to adjacent areas. 
These visualizations validate FengWu's ability to estimate weather states approximating the real data. 

\subsection{Effects of the Replay Buffer}
\begin{figure}
  \centering      
    \includegraphics[width=0.98\textwidth]{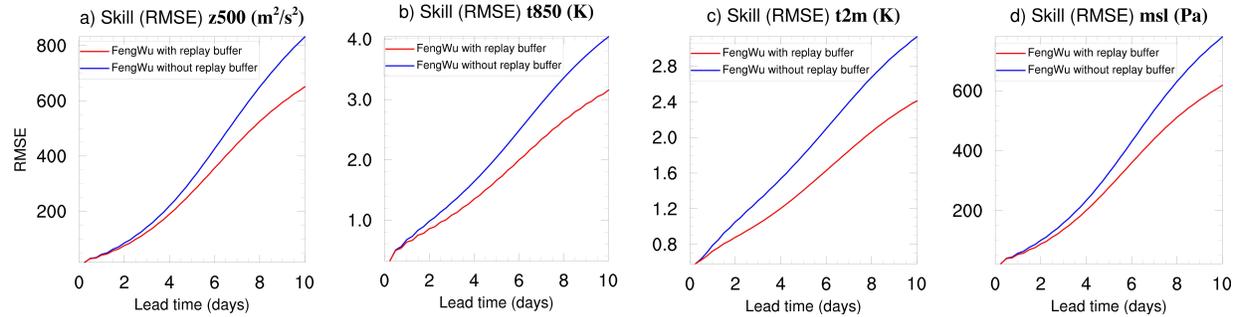}
    \caption{\textbf{Effects of the autoregressive training with the proposed replay buffer mechanism (red lines) or without it {(blue lines).} } 
    The x-axis in each sub-figure represents the lead time at 6-hour steps over 10 days. The y-axis represents RMSE defined in Eq.~\ref{eq:MSE} (lower is better).}
    \label{fig:ablation}
\end{figure}

As mentioned in Section~\ref{Sec:relpay_buffer}, the primary goal of the replay buffer is to reduce the accumulated error in the long-term prediction caused by the autoregressive estimation paradigm. To evaluate the effectiveness of the replay buffer in boosting forecasting skills, we compare the performance of FengWu with and without the replay buffer mechanism. As shown in Figure~\ref{fig:ablation}, the system's forecast performance decreases significantly as lead time increasing when the proposed replay buffer is removed, indicating that it is a crucial component in improving the accuracy of long-lead weather predictions.  

\subsection{Computation Cost}

\textbf{Training Cost}: FengWu is developed with Pytorch and trained with 32 Nvidia A100 GPUs in a cluster for 17 days in total. Compared with GraphCast trained with 32 Cloud TPU v4 for 21 days, we consider the training time of FengWu is only 47\% to 67\% of GraphCast (according to ~\cite{tpuv4}, TPU v4 is 1.2 to 1.7 times faster than Nvidia A100).

\textbf{Inference Cost}: We evaluate the inference speed of FengWu on an NVIDIA Tesla-A100 GPU, which indicates that FengWu costs less than 30 seconds to generate all forecasts in the following 10 days with a six-hour interval. With a peak power consumption of 0.4KW for an A100~\citep{choquette2021nvidia}, a 10-day inference by FengWu consumes roughly 12kJ energy, while the consumption of a single member of the IFS model is estimated to be about 26.6MJ \footnote{82 minutes is required by the IFS model in a 15-day, 51-member ensemble forecast applying 18km resolution grid  on 1530 Cray XC40 nodes with dual-socket Intel Haswell processors  ~\citep{bauer2020ecmwf}. A dual-socket Intel Haswell node draws a Thermal Design Power (TDP) of 270 Watts ~\citep{pathak2022fourcastnet}. The 10-day forecasts for a single IFS member roughly take 98,400 node seconds.},
approximately 2000 times higher than FengWu.

\section{Conclusions and Discussions} \label{sec:discussions}
 In this paper, we introduce FengWu, an advanced AI-based weather forecasting system, which has three technical contributions. 
%techniques
First, we propose to solve the problem of global medium-range weather forecasting as a multi-modal multi-task learning problem and introduce corresponding techniques for it, including multi-modal network architecture and uncertainty-based multi-task loss.
Second, we propose a replay buffer mechanism, which could improve the long-term forecasting performance under the autoregressive inference setting with limited devices. 
Third, FengWu achieves top performance among all existing AI-based methods and extends the skillful global medium-range weather forecast lead time to 10.75 days (ACC of z500 > 60\%). 
It also has higher accuracy than GraphCast, the current state-of-art AI-based weather forecasting model, on 80\% reported prediction targets.

As the initial fields of uneven quality are applied in physics-based and AI-based models, the fairness of model comparisons is supposed to be discussed. 
Specifically, the precision of weather forecasting systems greatly depends on the initial fields because the weather system is chaotic, and the chaotic system obtains precise estimations of its future with accurate initial values.
AI methods including FengWu and GraphCast ~\citep{lam2022graphcast} forecast weather with initial states from ERA5, while physics systems like IFS-HRES use their own initial analysis. 
To achieve timely forecasts, IFS-HRES produces such analysis fields with observations accessible at the start moment of prediction. For example, if observations from some satellites take several hours to be synchronized, they might be excluded from data assimilation for initial analysis states. Compared with IFS-HRES, ERA5 provides a more complete and accurate picture of initial fields, as the 5 days delay in ERA5's analysis data allows all observations to be blended. 
As a consequence, AI methods occupy a favorable position, for ERA5's premium initial states, in comparisons with IFS-HRES, while the comparison between AI methods, i.e., FengWu and GraphCast, is reasonable because their initial fields are both ERA5 reanalysis data.

\section*{Acknowledgements}
We acknowledge the use of the ERA5 dataset on both pressure levels and single level provided by the European Centre for Medium-Range Weather Forecasts (ECMWF). Without their great efforts in collecting, archiving, and disseminating the data, this study would not be possible. 

We acknowledge the Research Support, IT, and Infrastructure team based in the Shanghai AI Laboratory for their provision of computation resources and network support. F Ling and J-J Luo are supported by National Key Research and Development Program of China (No. 2020YFA0608000). This acknowledgment extends particularly to the individuals on the team, namely Prof. Yu Qiao, Liang Liu, Qihong Liao, Jiamin Ge, Jing Zou, Jingwen Li, and Xingpu Li.

We would also like to express our appreciation to Prof. Yang Wang from the University of Science and Technology of China, Prof. Tao Chen from Fudan University, Prof. Hongsheng Li from The Chinese University of Hong Kong, Dr. Jiajun Deng from the University of Sydney, Dr. Tong He from Shanghai AI Laboratory, and Mr. Peng Ye for their help and valuable discussions during the conduction of this research, which have significantly enhanced the quality of this work. We are grateful for their contributions and acknowledge their role in the development of this work.

\bibliographystyle{unsrtnat}
\bibliography{egbib}

\clearpage

%\section*{Appendix A: if needed}

%\clearpage

\end{document}